\documentclass[conference]{IEEEtran}
\IEEEoverridecommandlockouts
\usepackage{amsmath}
\usepackage{amssymb}
\usepackage{cite}
\usepackage{amsmath,amssymb,amsfonts}
\usepackage{algorithmic}
\usepackage{graphicx}
\usepackage{textcomp}
\usepackage{xcolor}
\usepackage{float}
\usepackage{tikz}
\usepackage{hyperref}
\usepackage{multirow}
\usepackage{booktabs} % For professional looking horizontal lines
\usepackage{array}
\def\BibTeX{{\rm B\kern-.05em{\sc i\kern-.025em b}\kern-.08em
    T\kern-.1667em\lower.7ex\hbox{E}\kern-.125emX}}
\begin{document}

\title{A 6G Integrated Sensing and Communication Framework for Railway Intrusion Detection and Collision Prediction}

\author{
    \IEEEauthorblockN{
        \small 
        Ajeet Kumar Yadav$^{1}$, 
        Sankaran Balasubramaniam$^{2}$, 
        Aritra Chatterjee$^{2}$, \\ 
        Vinod Aduru$^{2}$, 
        Yogesh Simmhan$^{3}$, and 
        Pandarasamy Arjunan$^{1}$
    }
    \IEEEauthorblockA{
        \small
        $^{1}$Department of Cyber Physical Systems, Indian Institute of Science, Bengaluru, India \\
        $^{2}$Nokia Solutions and Networks, Bengaluru, India \\
        $^{3}$Department of Computational and Data Sciences, Indian Institute of Science, Bengaluru, India \\
        \footnotesize 
        Emails: \{ajeety, samy, simmhan\}@iisc.ac.in, \\
        \{sankaran.balasubramaniam, aritra.chatterjee, vinod.aduru\}@nokia.com
    }
}

\maketitle

\begin{abstract}
Integrated Sensing and Communication (ISAC) refers to the combined use of sensing and communication systems to utilize wireless resources efficiently and has emerged as an essential paradigm for next-generation wireless networks. ISAC technology leverages wide bandwidth at high frequencies and a massive antenna array, both characteristic of 5G-advanced and 6G systems, and can perform sensing at the physical layer via Channel State Information (CSI). The 3rd Generation Partnership Project (3GPP) Technical Specification Group, in its Release 19 report, listed 32 potential use cases for ISAC, with a focus on detecting and tracking moving objects. In our work, we have addressed the Sensing for Railway Intrusion Detection use case. An intruder crossing the path of a running train can cause a major crash, especially when wildlife is involved. We have generated 22,695 CSI matrices and their ground truth using a three-dimensional (3D) rendered scene that replicates the real-world railway track and the Sionna radio simulator. We have also developed a three-dimensional Convolution Neural Network (3D CNN) and a Bidirectional Long Short-Term Memory (BiLSTM) based Machine Learning (ML) model to predict the presence of an intruder in the danger zone of the train track, and the intruder's real-time position relative to the train, velocity, and time of collision. On synthetic CSI data generated through simulation, for intruder detection, we achieved 99.57\% classification accuracy on a balanced test data, and for position, velocity, and time-of-collision prediction, a combined Mean Absolute Error (MAE) of 0.4240. These results demonstrate the potential of CSI-based ISAC sensing combined with ML for reliable railway intrusion detection. The complete codebase for CSI data generation, preprocessing, and model development is publicly available here \href{https://github.com/EdgeIntelligenceLab/6g-isac-railway-intrusion-detection}{https://github.com/EdgeIntelligenceLab/6g-isac-railway-intrusion-detection}.
\end{abstract}


\section{Introduction}
The evolution of wireless networks from communication-centric systems toward intelligent networks capable of perceiving their physical surroundings has positioned ISAC as a key enabling technology for 5G-Advanced and future 6G systems \cite{isac}. Recent advances in cellular networks have expanded their capabilities toward a wide range of sensing applications \cite{isac_uses}. By integrating wireless communication and sensing into a unified radio interface and control framework, ISAC enables cellular infrastructure to provide robust connectivity and real-time environmental awareness, including object detection, localization, and monitoring \cite{wifi_vision}.
\begin{figure}[!t]
    \centering
    \includegraphics[width=1\linewidth]{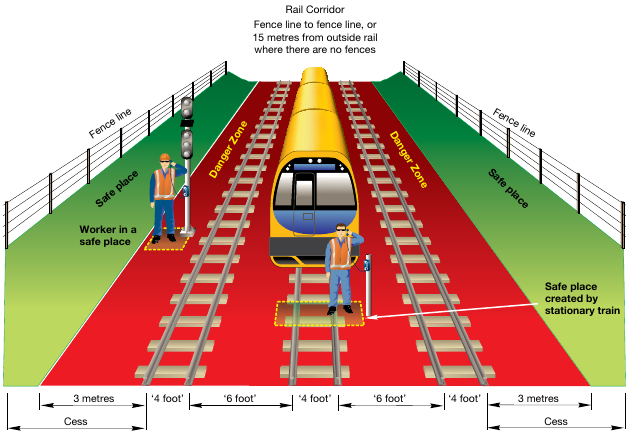} 
    \caption{An illustration of the rail corridor, danger zone, and safe places. The danger zone is defined as the region within 3 m horizontally from the nearest track. Source: Rail Industry Safety Induction Handbook, Sydney Trains \cite{rail_safe}.}
    \label{fig:rail_safe}
\end{figure}

Recognizing ISAC's broad potential, the 3GPP Release 19 identified 32 use cases, among which \emph{railway intrusion detection} is considered an important application for improving railway safety~\cite{3gpp}. Railway networks span thousands of kilometers across diverse environments, including urban areas, forests, mountains, and rural regions, where maintaining continuous situational awareness through surveillance is particularly challenging \cite{challenges}. Unpredictable conditions make foreign object intrusion a major threat to railway safety, particularly when pedestrians, falling rocks, wildlife, and other objects are present within the railway danger zone, which is illustrated in Figure \ref{fig:rail_safe}. Many railway tracks are equipped with fencing; however, this is often insufficient to prevent wildlife intrusion, especially in remote areas \cite{elephant}. This use case motivates the investigation of whether communication-derived radio measurements can provide sufficient information not only to detect an intrusion but also to characterize the intruder's motion and collision risk.

Various sensing technologies have been investigated for railway intrusion detection. Infrared light spot-based intruder detection systems are precise but have a very limited range and may be suitable for indoor detection \cite{BriGuard}. They depend heavily on environmental factors and the temperature of objects. LiDAR-based methods are designed for fixed areas with high-risk factors rather than for covering long sections of the rail track \cite{LiDAR, 3D_LiDAR}. Their detection rate is distance-dependent and decreases rapidly with distance. Camera-based detection is very precise and can distinguish between animals, humans, and vehicles, but the camera's field of view is limited, and if the intruder is not in Line of Sight (LoS), detection becomes difficult \cite{MACENet, Animal_detection, BCD-YOLO}. Additionally, their performance is highly dependent on lighting conditions and sensitive to environmental variations. Despite extensive research on vision-, LiDAR-, and sensor-based railway intrusion detection, the potential of CSI-based \cite{csi} wireless sensing for addressing the 3GPP railway intrusion detection use case remains largely unexplored. In particular, existing approaches primarily focus on obstacle detection or classification, while jointly estimating motion-related parameters such as relative position, velocity, and collision timing directly from high-dimensional CSI remains an open research problem. Furthermore, obtaining sufficiently diverse labeled CSI measurements for such safety-critical railway scenarios is challenging, motivating physics-based simulation and data generation. 

CSI-based wireless sensing provides a practical approach for realizing ISAC by exploiting variations in the wireless propagation channel induced by objects and their movements in the surrounding environment \cite{isac_csi}. The presence and motion of objects alter the multi-path propagation environment through changes in reflection, scattering, diffraction, and Doppler characteristics. These variations are encoded in CSI across antenna, frequency, and temporal dimensions. Consequently, a sequence of CSI measurements can capture not only whether an intruder is present but also how its position and velocity vary. However, extracting meaningful information from CSI measurements remains challenging due to the high dimensionality, noise, and temporal dynamics of wireless channels. Recent advances in machine learning provide an effective means of learning discriminative spatio-temporal representations directly from CSI, making machine learning a key enabler for Intelligent 6G sensing applications \cite{csi_ml}. 

In this work, we address the 3GPP-defined railway intruder detection use case through a CSI-based sensing framework integrated with machine learning. We render a high-fidelity three-dimensional railway platform and generate synthetic CSI measurements using the Sionna \cite{sionna} radio simulator based on realistic wireless propagation. A total of 22,695 CSI matrices with corresponding ground-truth labels are generated for ML model training and evaluation. The raw CSI matrices are then processed using a signal-processing pipeline comprising subcarrier selection, static component removal, and temporal denoising. Finally, a hybrid machine learning architecture combining 3D CNN \cite{3d_cnn} and BiLSTM \cite{lstm} networks is developed to learn spatio-temporal representations from CSI matrices to jointly predict the presence of an intruder, its position, velocity, and time-to-collision. 
\begin{figure*}[!t]
    \centering
    \includegraphics[width=1\linewidth]{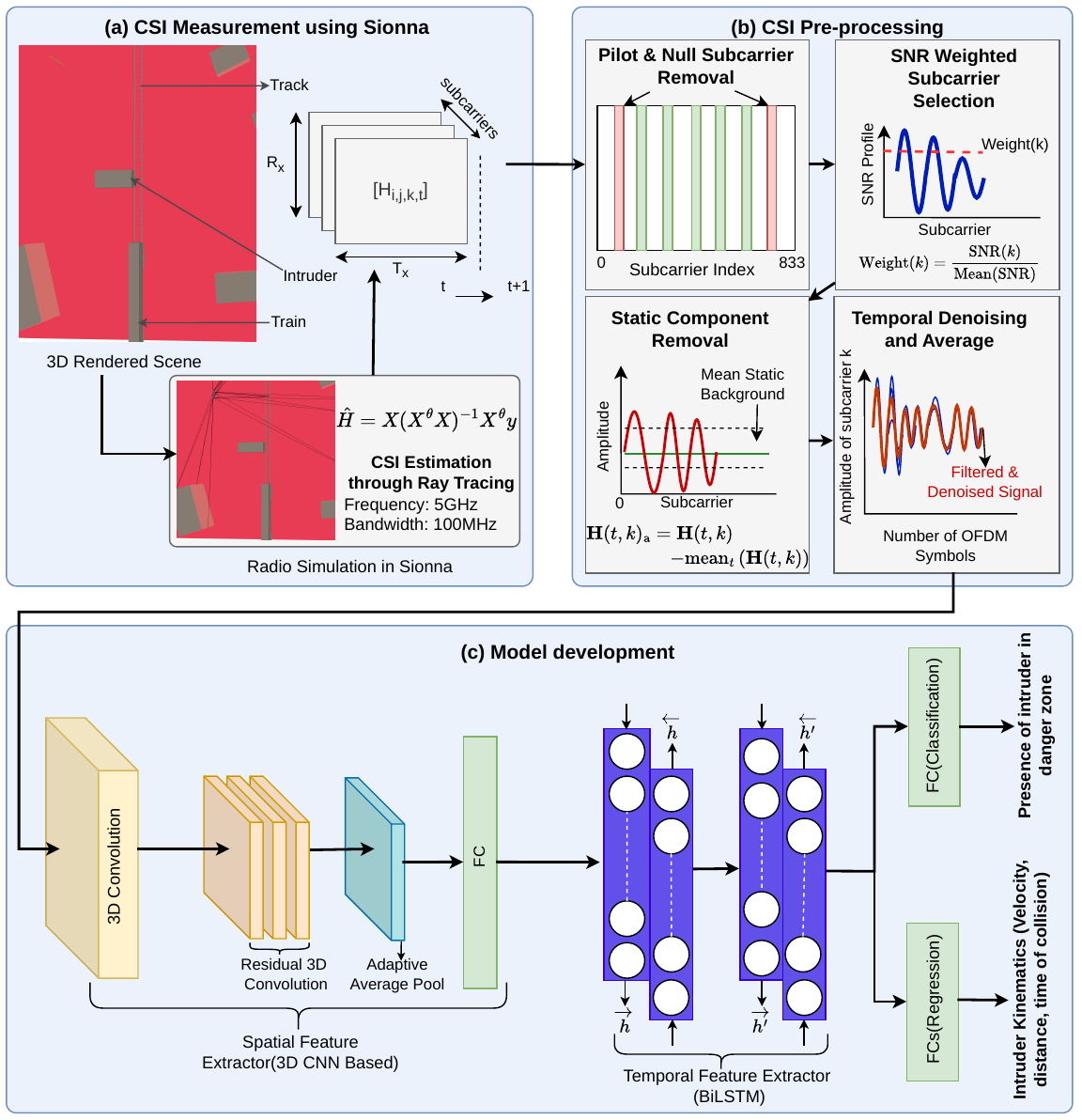} 
    \caption{A visual description of our complete approach. First, we generate CSI \cite{csi} data using the Sionna \cite{sionna} simulation. After that, we pre-process raw CSI data using Pilot \& Null Subcarrier Removal, Signal to Noise Ratio (SNR) Weighted Subcarrier Selection, Static Component Removal, and Temporal Denoising methods. Once the data is ready and fed to the ML Model for training, the trained model predicts the presence and kinematic parameters of the intruder.}
    \label{fig:overview}
\end{figure*}

Extensive simulation and evaluation demonstrate the method's effectiveness for sensing in railway environments. On the generated CSI dataset, the proposed model achieves an intrusion detection accuracy of 99.57\% on a balanced test set while simultaneously estimating the intruder's position, velocity, and time-to-collision with a combined MAE of 0.4240. These results demonstrate the potential of CSI-based sensing and machine learning for intelligent railway safety applications in future ISAC networks.

The major contributions of this work are as follows:
\begin{itemize}
\item We develop a CSI-based wireless sensing pipeline for the 3GPP-defined railway intrusion detection use case, extending conventional intrusion detection beyond binary object presence to jointly estimate the intruder's relative position, velocity, and time-to-collision to assess collision risk.
\item We use a physics-based CSI data generation pipeline using a high-fidelity three-dimensional railway environment and the Sionna radio simulator. The pipeline generates CSI matrices with corresponding ground-truth intrusion and kinematic information under varying railway scenarios, providing labeled data for training and evaluating ML models.
\item We design an end-to-end CSI processing and machine learning model that combines subcarrier selection, static component removal, and temporal denoising with a hybrid 3D CNN--BiLSTM architecture. 
\item We conduct extensive experiments under two wireless communication configurations and evaluate the proposed model using both classification and regression metrics. The proposed model's confidence score on intrusion detection and intruder's position, velocity, and time-to-collision demonstrates the potential of CSI-based sensing for intelligent railway safety applications.
\end{itemize}

The rest of this paper is structured as follows. Section II reviews the related work on ISAC and existing railway intruder detection methods. Section III presents the methodology of the proposed solution, including CSI data generation, exploratory data analysis, CSI preprocessing, the proposed 3D CNN–LSTM architecture, and the training procedure. Section IV presents the experimental results and evaluates the proposed framework under different validation settings. Finally, Section V concludes the paper and outlines limitations.
\section{Related works}
Beyond communication, 6G is also set to revolutionize sensing applications. Recent research has demonstrated that 6G can enable highly accurate localization and motion detection across a wide range of objects \cite{positioning}. Not much prior work integrates 6G-based sensing for real-time multi-parameter railway intruder characterization (intruder presence, velocity, position, minimum-distance locus, collision timing). This section reviews (i) foundational 6G ISAC architectures enabling sensing from communication signals, (ii) existing railway intrusion detection systems lacking 6G-scale kinematic estimation.
\subsection{6G Integrated Sensing and Communication}
ISAC is a cornerstone of 6G wireless networks, enabling the joint use of communication waveforms for high-resolution environmental sensing and connectivity. Traditional radar systems required dedicated spectrum and hardware; ISAC leverages Orthogonal Frequency-Division Multiplexing (OFDM) \cite{ofdm}- based communication signals to estimate presence, velocity, and position via CSI perturbations caused by environmental reflectors. Bayasteh et al. presented a practical ISAC deployment that achieves centimeter-level localization and multi-target tracking using mono-static, bistatic, and multi-static configurations \cite{bayesteh}. A Ghosh et al. provided an overview of different sensing topologies and outlined ISAC requirements across various deployment contexts \cite{ghosh}. They also discussed waveform design, beamforming strategies, and associated hardware requirements. Sebastian et al. presented multiple test cases of ISAC along with their architecture \cite{sebastian}. These studies collectively highlight the growing demand for ISAC.
\subsection{Railway Intrusion Detection Systems}
Railway intruder detection systems have grown from traditional track circuitry to multi-modal sensor fusion, driven by the need for higher detection ranges and collision avoidance in high-speed rail networks. Zongliang et al. presented a high-precision obstacle-detection algorithm using a 3D mechanical LiDAR to meet railway safety requirements \cite{LiDAR}. This method can precisely detect the presence of small objects on the rail track. In addition, significant work has been done on intruder detection using vision-based ML models. Shanping et al. \cite{Shanping} proposed a railway intrusion and risk quantification algorithm using track semantic segmentation and spatiotemporal features, whereas in his work, Hu T \cite{Hu} used the MSIA-YOLOv8 and DALNet models to detect multi-scale obstacles. These provide detection but lack the 6G-sensing kinematic estimation needed for collision avoidance.

This work uniquely combines 6G CSI sensing with railway safety, predicting presence, minimum distance locus, velocity vectors, and collision timing from the same OFDM waveform used for communication.

\section{Methodology} 
The complete process of our approach is divided into three major stages, as shown in Figure \ref{fig:overview}. In the first stage, synthetic CSI data is generated via radio simulation, with the schematic designed to produce realistic, high-quality CSI-RS configurations for railway sensing applications. The setup includes three key components: a 3D environment model representing a rendered scene with multiple objects, a wireless channel model consisting of a transmitter and receiver, and physics-based ray-tracing steps for CSI extraction. The second stage preprocesses the raw CSI matrices into a precise, dense feature representation. In the third stage, a sufficiently strong ML model is developed to precisely learn and extract features from the synthetic CSI data and map them to the corresponding outputs through training. Finally, the trained model is validated on unseen CSI data to determine its performance confidence score. Next, the data generation and the ML model architecture description are given in detail.
\begin{figure}
    \centering
    \includegraphics[width=1\linewidth]{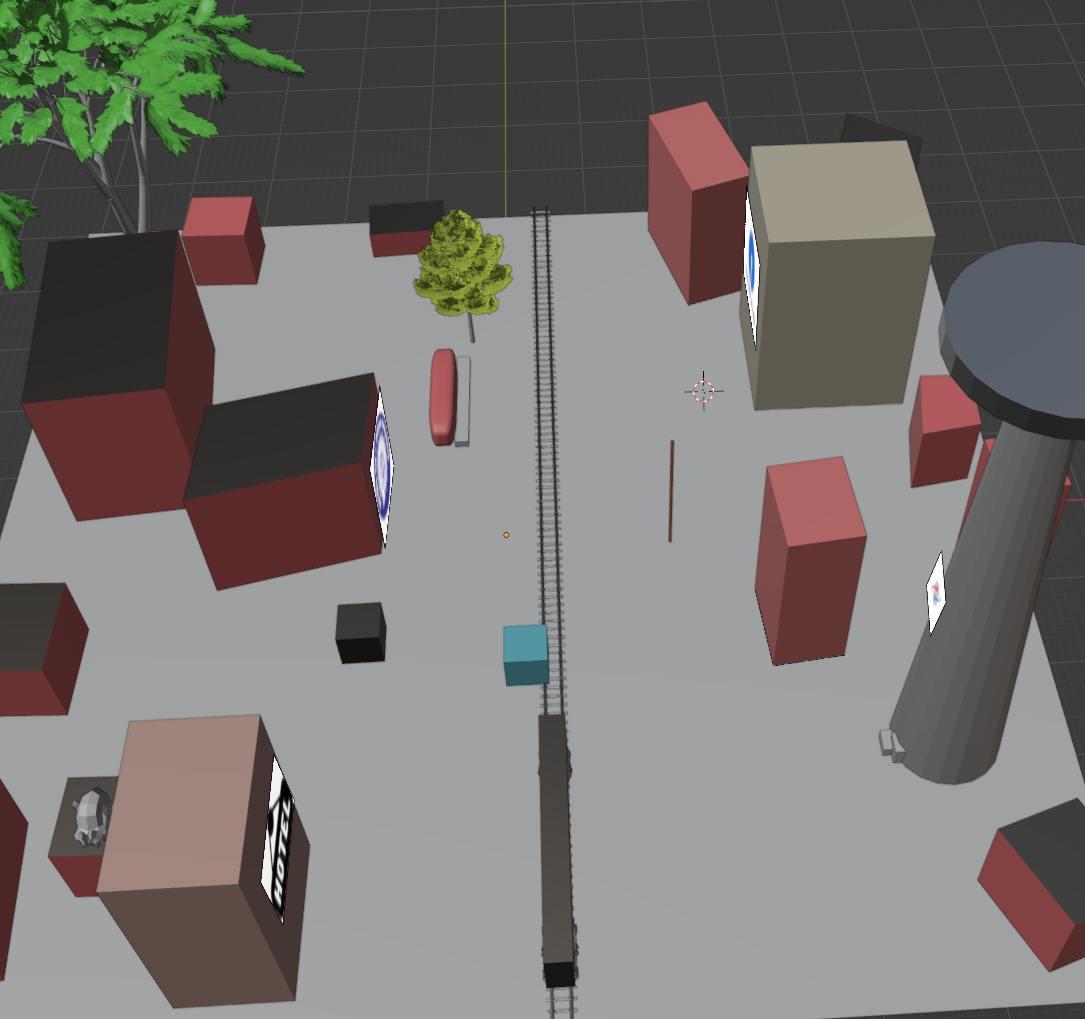}
    \caption{A 3D rendered scene of a train platform with the help of Blender \cite{blender}. This scene contains multiple concrete buildings, trees, and a tower, some of which are fully reflective, while others are partially reflective. The train track is in the middle of the scene along the y-axis, and the train is at the bottom of the scene, along with a sky-blue intruder that can move on the platform.}
    \label{3D Scene}
\end{figure}
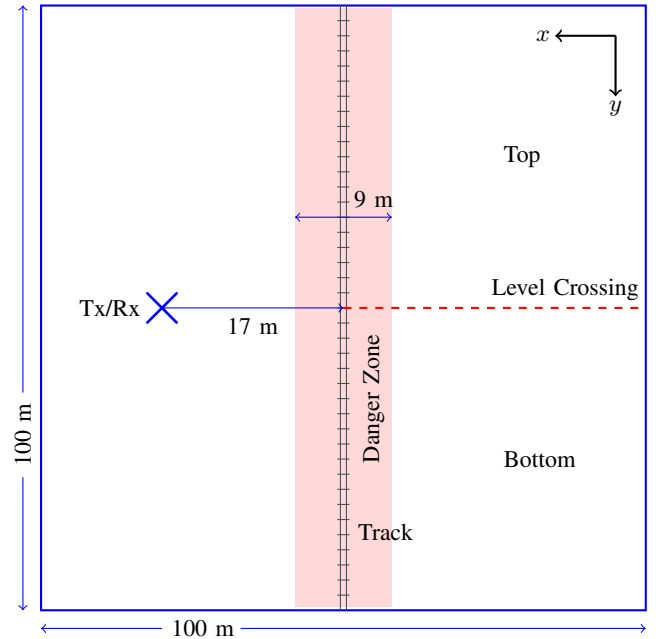
\begin{figure}
\centering
\begin{tikzpicture}[scale=0.8, every node/.style={font=\small}]
% Outer border
\draw[blue, thick] (-5,-5) rectangle (5,5);
% Danger zone (shaded rectangle)
\fill[red!15] (-0.8,-4.95) rectangle (0.8,4.95);
% Track (vertical line)
\draw[black!70] (0.05,-5) -- (0.05,5);
\draw[black!70] (-0.05,-5) -- (-0.05,5);
\node[above] at (0.7,-4) {Track};
% Tick marks on track
\foreach \y in {-4.75, -4.5,...,4.75}
\draw[black!70] (-0.1,\y) -- (0.1,\y);
% Posistion of antenna
\draw[<->, blue] (0.0, -0.0) -- (-3,-0.0);
\node[below] at (-1.5, -0.0) {17 m};
% Level crossing line (dashed)
\draw[red, dashed, thick] (0,0) -- (5,0);
\node[rotate=90] at (0.5,-1.5) {Danger Zone};
% Width (9m)
\draw[<->, blue] (-0.8, 1.5) -- (0.8, 1.5);
\node[above] at (0.5, 1.5) {9 m};
% Tx/Rx position
\node[blue] at (-3,0) {\Huge $\times$};
\node[left] at (-3.2,0) {Tx/Rx};
% Distance 100 m
\draw[<-, blue] (-5.3,-5) -- (-5.3, -2.8);
\draw[->, blue] (-5.3,-1.4) -- (-5.3, 5);
\node[rotate =90] at (-5.3,-2.1) {100 m};
% Distance 100 m
\draw[<-, blue] (-5, -5.3) -- (-3, -5.3);
\draw[->, blue] (-1.7, -5.3) -- (5, -5.3);
\node[right] at (-3.0,-5.3) {100 m};

% Danger zone label
\node[right] at (2.3,0.3) {Level Crossing};
% Top and Below labels
\node[right] at (2.5,2.5) {Top};
\node[right] at (2.5,-2.5) {Bottom};

% ===== X-Y AXIS (Top Right Corner) =====
\begin{scope}[shift={(4.5,4.5)}]
    % +x to LEFT (label slightly above the line)
    \draw[->, thick] (0,0) -- (-1,0) node[pos=1.2] {$x$};    
    % +y DOWN (label slightly right of the line)
    \draw[->, thick] (0,0) -- (0,-1) node[pos=1.2] {$y$};
\end{scope}
\end{tikzpicture}
\caption{The 2D layout of the 3D-rendered scene is shown in the diagram. In this, the coordinate system origin (0, 0) is the centre of the plane. The convention of the x-axis and the y-axis is given in the top right portion of this plane. The train track is along the y-axis, and the danger zone is defined according to 3GPP \cite{3gpp}: 3 m on either side of the track, with a track width of 3 m. The antenna position is (17, 0), and the plane portion above the Level Crossing line is the top portion of the plane, and the plane portion below the Level Crossing line is the Bottom portion of the plane.}
\label{2D_scene}
\end{figure}
\begin{figure*}
    \centering
    \includegraphics[width=\linewidth, height=1\textheight, keepaspectratio]{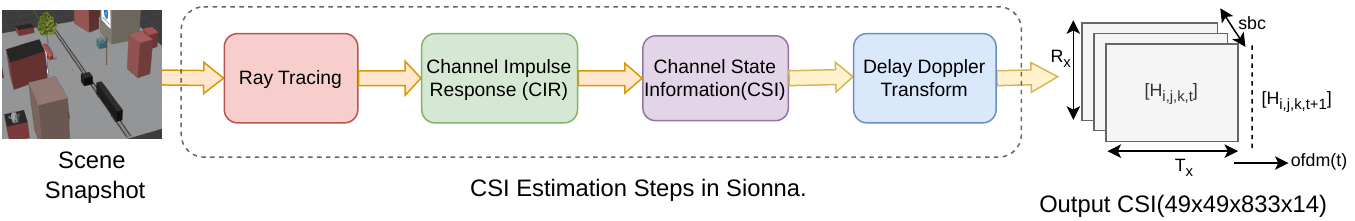}
    \caption{Physics-based CSI calculation procedure in Sionna. Although sensing is continuous, to generate the CSI, Sionna takes a snapshot of the scene and performs the calculation in four major steps: Ray Tracing, Channel Impulse Response, Channel State Information, and Delay-Doppler transform. In the output, we get a complex-valued matrix whose shape depends on the radio configuration. In this, the output with shape 49x49x833x14 is shown using a specific radio configuration defined in the upcoming section. In the output CSI shape, sbc refers to subcarriers.}
    \label{csi_cal}
\end{figure*}
\subsection{CSI Data Generation}
To create a realistic railway environment, a 3D scene is modeled in Blender \cite{blender} by integrating a variety of materials with varying reflective properties replicate a real-world setting, as illustrated in Figure \ref{3D Scene}. The core of the environment features several reflective metallic components, specifically a metal train stationed alongside a metal track and a tall metallic tower. These surfaces are placed to interact significantly with radio signals. The surrounding infrastructure consists of a few marble buildings and several concrete buildings, all treated as partially reflective surfaces to simulate natural stone and masonry finishes. Adding to the scene's complexity are organic elements, such as two trees with partially reflective stems.  Finally, a reflective metal surface intruder is placed and can move in any direction on the platform. 

The 2D spatial layout of the same 3D scene is defined by a square platform measuring 100 m by 100 m, resulting in a total area of 10,000 m², which can be observed in the Figure \ref{2D_scene}. The coordinate system is centered at (0, 0), which marks the specific intersection where the railway track meets the level crossing line. Within this, the antenna is positioned at (17 m, 0). The track itself extends from (0, -50 m) to (0, 50 m), having a width of 3 m, with a designated danger area spanning 3 m on either side of the rails as shown in Figure \ref{fig:rail_safe} and proposed by \cite{3gpp}. The regions located above and below the level crossing line are categorized simply as the top and bottom areas of the scene, respectively. CSI is subsequently estimated using the Sionna-based radio simulation.

\subsubsection{CSI Estimation}
For sensing simulation, we have used Sionna, a physical-layer simulation library developed by NVIDIA. We have used the Sionna 0.19.2 version to perform simulations. The rendered 3D scene is integrated with Sionna's ray tracing and channel modeling pipeline. Rays are traced between the transmitter and the receiver to simulate realistic multi-path propagation. The simulation output consists of CSI matrices that capture the characteristics of the communication channel and the impact of obstructing objects across the defined OFDM subcarriers and antenna configuration.
This process simulates electromagnetic signal propagation between a Transmitter (Tx) and a Receiver (Rx), both mounted on the static antenna pole defined in the layout. By tracing the paths of the rays as they interact with the various reflective and partially reflective surfaces in the environment, the program generates raw data regarding signal behavior. Based on these results, multiple calculations are performed to derive the CSI. The estimation of the CSI is a multi-step procedure, which is depicted in Figure \ref{csi_cal}.

To understand the CSI estimation process in Sionna in a concrete way, we follow a few conventions.
\begin{itemize}
\item \textbf{A scene}: A 3D rendered space where the dimensions of all objects are fixed. Objects can start moving from any position with any velocity.
\item \textbf{A scene setup}: A 3D rendered scene where the velocity and initial position of the movables are fixed. They can move from a defined point with a defined velocity.
\item \textbf{A simulation}: One complete run of a scene setup in Sionna is called one simulation. A complete simulation generates many CSI matrices.
\item \textbf{Step}: Part of the simulation in which one CSI matrix gets generated is called a step. A completed simulation of a scene setup is carried out in multiple steps, and hence, it generates many CSI matrices. We conducted multiple simulations with up to 45 steps, and several extended simulations with up to 120 steps.
\item \textbf{Scene Snapshot}: A scene snapshot is a diagram of a 3D scene setup at a particular step. We take a scene snapshot for better visualization of the movement of the object. A scene snapshot contains the moving and non-moving object status at a step when the initial position and velocities are fixed on moving objects. The receiver receives reflected and refracted rays from various objects at that step.
\item \textbf{CSI Snapshot}: A CSI snapshot is a record of the CSI matrix at a step, and we call it a CSI matrix.
\end{itemize}

Next, we outline the step-by-step procedure for calculating the CSI matrix within the Sionna, as illustrated in Figure \ref{csi_cal}.

\textbf{\textit{Ray Tracing}}:
Rays that travel in a straight line show the reflection, refraction, diffraction, and scattering phenomena due to obstacles in their path. Ray tracing is a method for determining the path of a ray from its source to an object or receiver. A traced ray may directly reach the receiver or after multiple reflections and refractions. Sionna uses the differentiation method for ray tracing, which helps determine material properties, gradients, antenna array patterns, and other aspects \cite{ray_trace}.

\textbf{\textit{Channel Impulse Response (CIR)}}:
CIR is like a fingerprint of the channel. It tells how an impulse sent through the channel spreads in time due to reflections and refractions, and conveys when and how strong each reflected path of the signal arrives at the receiver. CIR is a time-domain-based representation. The received signal is obtained by the convolution of the CIR with the transmitted signal. The time domain channel impulse response between transmitter and receiver can be written-
\begin{equation}
h(t) = \sum_{p=1}^{P} \alpha_p \, \delta(t - \tau_p)
\label{eq:channel_impulse_response}
\end{equation}
where $h(t)$ denotes the channel impulse response as a function of time, 
$P$ represents the total number of propagation paths, 
$\alpha_p$ is the complex gain associated with the $p^{th}$ path, 
$\tau_p$ denotes the propagation delay of the $p^{th}$ path, 
and $\delta(\cdot)$ represents the Dirac delta function.

\begin{figure}[!t]
    \centering
    \includegraphics[width=1\linewidth, height=0.2\textheight, keepaspectratio]{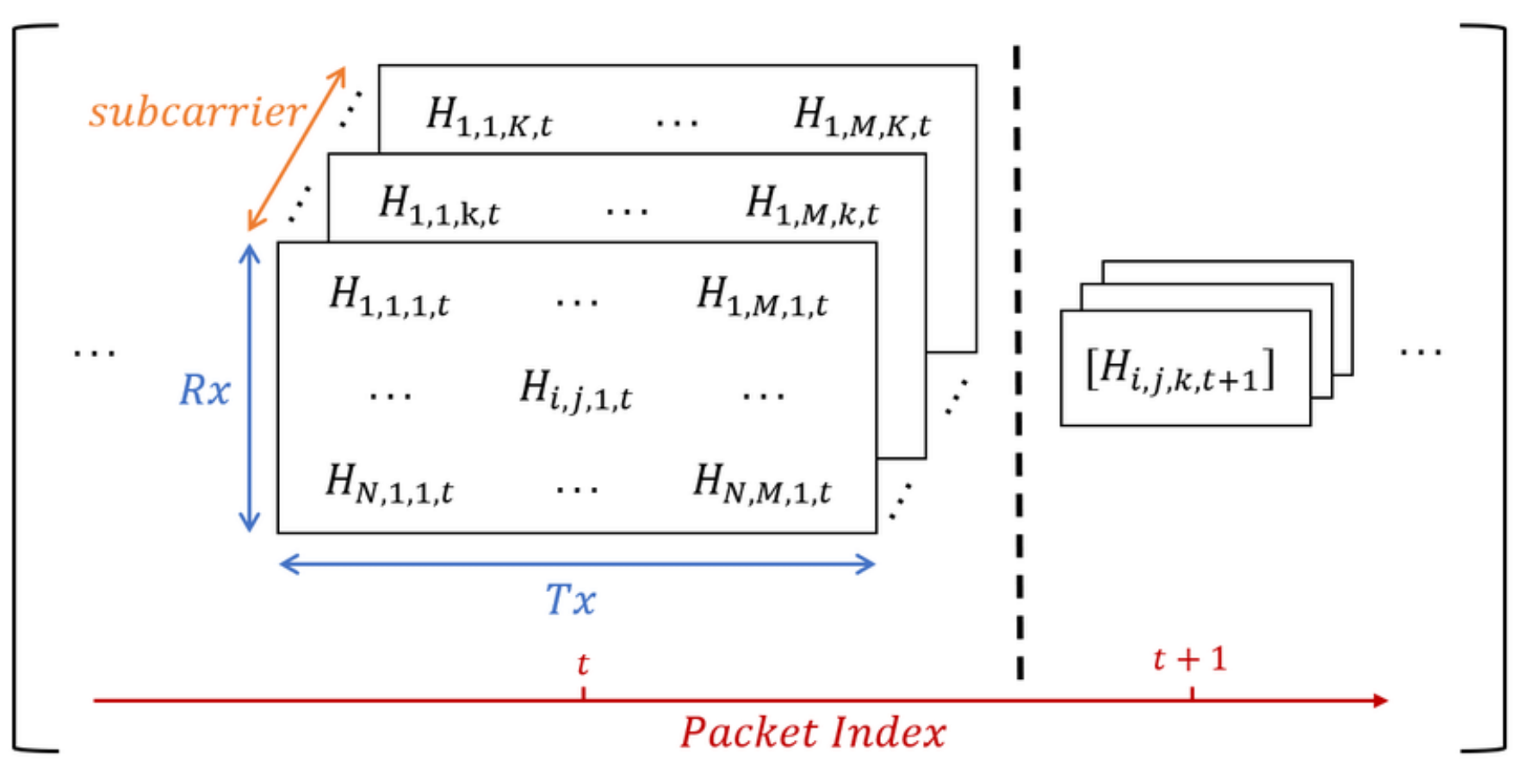}
    \caption{Shape of a standard CSI matrix as given by \cite{wireless_sensing} across Tx and Rx dimensions over packet indices $t$ and $t+1$. Each packet index contains an NxMxK cube, where $N$, $M$, and $K$ correspond to the transmit antenna grid size, the receive antenna grid size, and the subcarriers, respectively.}
    \label{csi_shape}
\end{figure}

\textbf{\textit{Channel State Information (CSI)}}:
CSI characterizes the impact of the wireless channel on the transmitted signal, capturing the effects of path loss, multi-path propagation, fading, and scattering. It is represented as a complex-valued matrix across antennas and subcarriers in an OFDM system. CSI provides a detailed description of channel conditions at a given time, enabling advanced tasks such as localization and sensing. As objects in the scene move, the multi-path structure changes, resulting in time-varying CSI snapshots. The shape of the CSI matrix depends on the number of Tx, Rx, subcarriers, and the number of OFDM symbols, as shown in Figure \ref{csi_shape}. CSI can be evaluated as-
\begin{equation}
H[k,n] = \sum_{p=1}^{P} \alpha_p 
e^{-j2\pi f_k \tau_p}
\cdot
e^{j2\pi f_{d,p} n T_s}
\label{eq:csi_model}
\end{equation}
where $H[k,n]$ denotes the CSI matrix at subcarrier $k$ and OFDM symbol $n$. 
$P$ represents the total number of propagation paths. 
$\alpha_p$ is the complex gain associated with the $p^{th}$ path. 
$f_k$ denotes the frequency of the $k^{th}$ subcarrier given by

\begin{equation}
f_k = f_c + k\Delta f
\end{equation}
where $f_c$ is the carrier frequency and $\Delta f$ is the subcarrier spacing. 
$T_s$ denotes the OFDM symbol duration, $\tau_p$ represents the propagation delay of the $p^{th}$ path, and $f_{d,p}$ denotes the Doppler shift corresponding to the $p^{th}$ propagation path.

\textbf{\textit{Delay Doppler transform}}:
Since our work involves moving objects, Doppler effects are very likely to occur. Since the scenario involves moving objects, Doppler shifts naturally arise in the received signal and provide useful information for estimating target motion. The Delay-Doppler Transform converts the time-domain channel impulse response into a 2D representation, with the vertical axis representing propagation delay and the horizontal axis representing Doppler frequency shift. This transformation is crucial for 6G sensing because
it separates and reveals two fundamental characteristics of moving objects: their spatial location and their velocity. It can be calculated as:

\begin{equation}
H(\tau, f_d) =
\int_{-\infty}^{\infty}
h(t,\tau)\, e^{-j2\pi f_d t} \, dt
\label{eq:delay_doppler}
\end{equation}
where $H(\tau, f_d)$ represents the two–dimensional delay–Doppler channel representation. $\tau$ denotes the propagation delay, and $f_d$ represents the Doppler frequency shift. 
The function $h(t,\tau)$ denotes the time-varying channel impulse response.

\begin{table}[!t]
\centering
 \caption{The variables used for data generation, along with their corresponding ranges and variation intervals used to cover these ranges, are listed in this table}
 \label{table:variable}
  \begin{tabular}{ | m{6em}  m{1.7cm} m{1.6cm}  m{1.8cm} |}
    \hline
    \textbf{Variable} & \textbf{Minimum Value} & \textbf{Maximum Value} & \textbf{Change} \\
    \hline
    Initial Position of Intruder & [-20, -21, 0]m & [10, 20, 0]m & [1, 1, 0] \\
    \hline
    Velocity of Intruder & [-1, 0, 0]m/sec & [1, 0, 0]m/sec & [1, 0, 0] \\
    \hline
    Train Velocity & 10m/sec & 10m/sec & 0 \\
    \hline
    Intruder Geometry & 1.5m×0.5m×1m & 10m×4m×5.5m & 0.5mx0.5mx0.5m\\
    \hline
  \end{tabular}
\end{table}
\subsubsection{Variables in CSI Estimation and Procedure}
To train the machine learning model, sufficient data is required. A single simulation can generate only a few CSI matrices, but multiple simulations can generate sufficient data to train the ML model. In the data generation phase, the simulation incorporates several variables to span the entire scene platform by moving the intruder and train. The train velocity is held constant at 10 m/s moving in the $-y$ direction (toward the top side or the negative y-axis). The primary variables involve the intruder, whose initial position, dimension and velocity are changed to test different detection scenarios. The intruder's velocity magnitude ranges from 0 to 1 m/s, with movement directions spanning the 2D plane, including left-to-right, right-to-left, and diagonal paths. Furthermore, the intruder's size varies from a minimum of 1.5 m × 0.5 m × 1 m, representing a difficult-to-detect target, to a maximum of 10 m × 4 m × 5.5 m, which serves as the standard case for most simulations. Table \ref{table:variable} lists the variables and their ranges. By varying these variables, we can generate a large number of data samples to train the ML model. Out of all variables given in Table \ref{table:variable}, the most varied variable is the initial position of the intruder, and the least varied variable is the intruder geometry. In 95\% of simulations, CSI data is recorded at the intruder's maximum dimension.
\begin{figure*}[!t]
    \centering
    \includegraphics[width=1\linewidth, height=0.5\textheight, keepaspectratio]{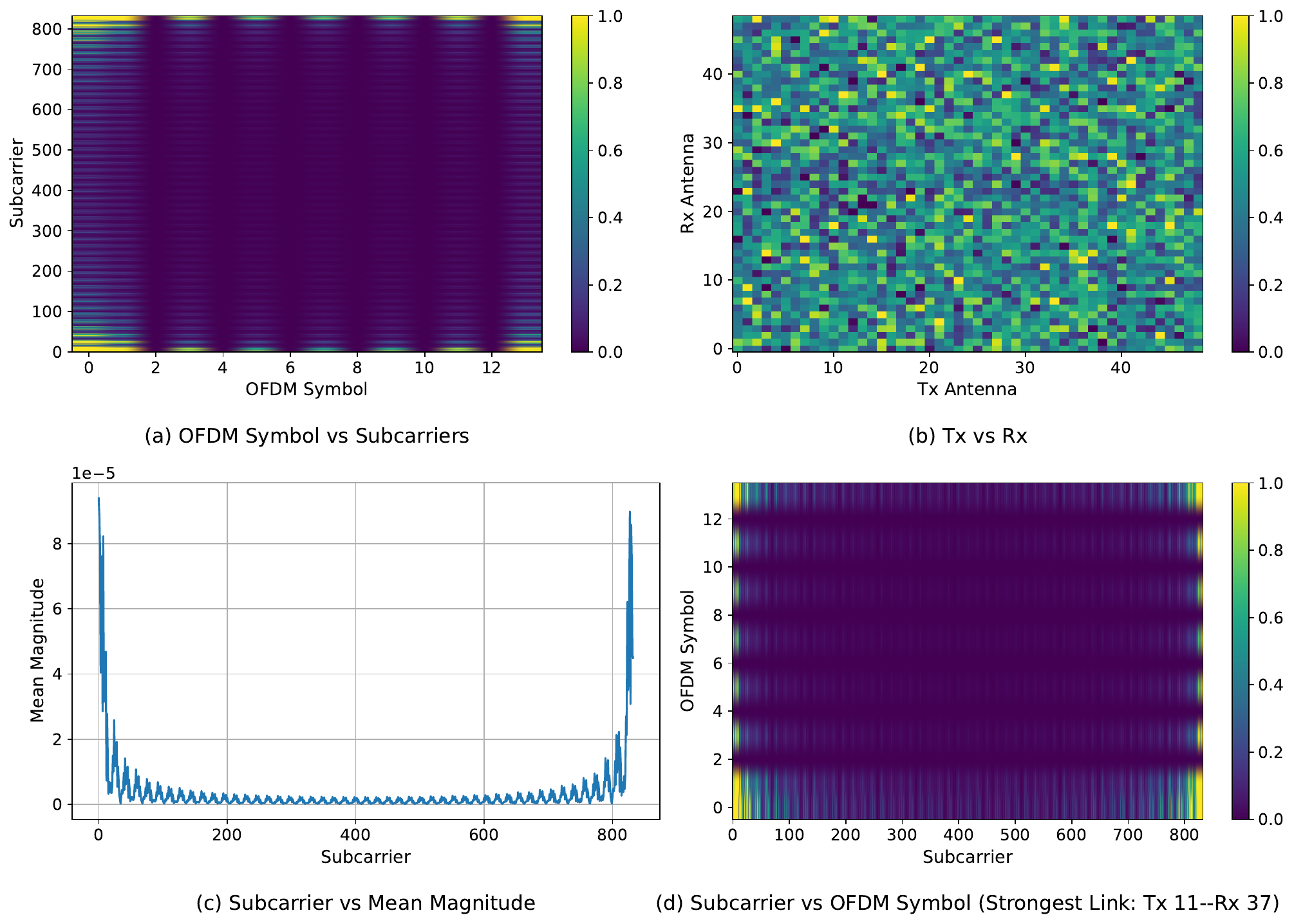}
    \caption{CSI magnitude visualisation using a four-step method when an intruder is present in the danger zone. Current position of moving intruder is [-5.79, -10] m, and distance from intruder edge to train edge is 5.9 m. This complete visualisation is plotted in a 2x2 grid, out of which (a)  represents the Frequency-Time profile,  (b) tells about the MIMO Channel structure, (c) tells about the Signal Evolution over frequency, and (d) talks about the strongest MIMO link between transmitter and receiver.}
    \label{fig:csi_mag}
\end{figure*}
A simulation done through a systematic, step-by-step procedure within the Sionna. Initially, a single scene setup is loaded, fixing the objects' geometry and material properties. Once the train and the intruder velocity and position are initialised, the train and the intruder begin their synchronised movement while the static antenna transmits and receives signals. Although signal transmission is continuous, the CSI recording process occurs in discrete steps. During each step, the positions of the moving objects are updated, and Sionna generates a corresponding CSI matrix. This cycle repeats multiple steps per simulation to reach the final position of moving objects. Once they reach their destination, the next simulation can start by initializing new positions.

\subsection{Exploratory Data Analysis}
We analyzed the dataset and, after that, developed an appropriate ML model. The generated CSI matrices are multidimensional complex-valued matrices. The dimensionality of these matrices depends on the radio configuration. Radio configuration encompasses the tuned radio parameters in the simulation for Sionna. The standard shape of a CSI matrix is given in Figure \ref{csi_shape} as suggested by \cite{wireless_sensing}. In this CSI shape, Tx and Rx depend on the number of antennas in the transmitter and receiver, respectively. The number of subcarriers depends on the allotted bandwidth and subcarrier spacing. The fundamental block of the packet index is the number of OFDM symbols. Collectively, these radio parameters determine the dimensions of the resulting CSI matrix.
\begin{figure*}[!h]
    \centering
    \includegraphics[width=1\linewidth, height=0.5\textheight, keepaspectratio]{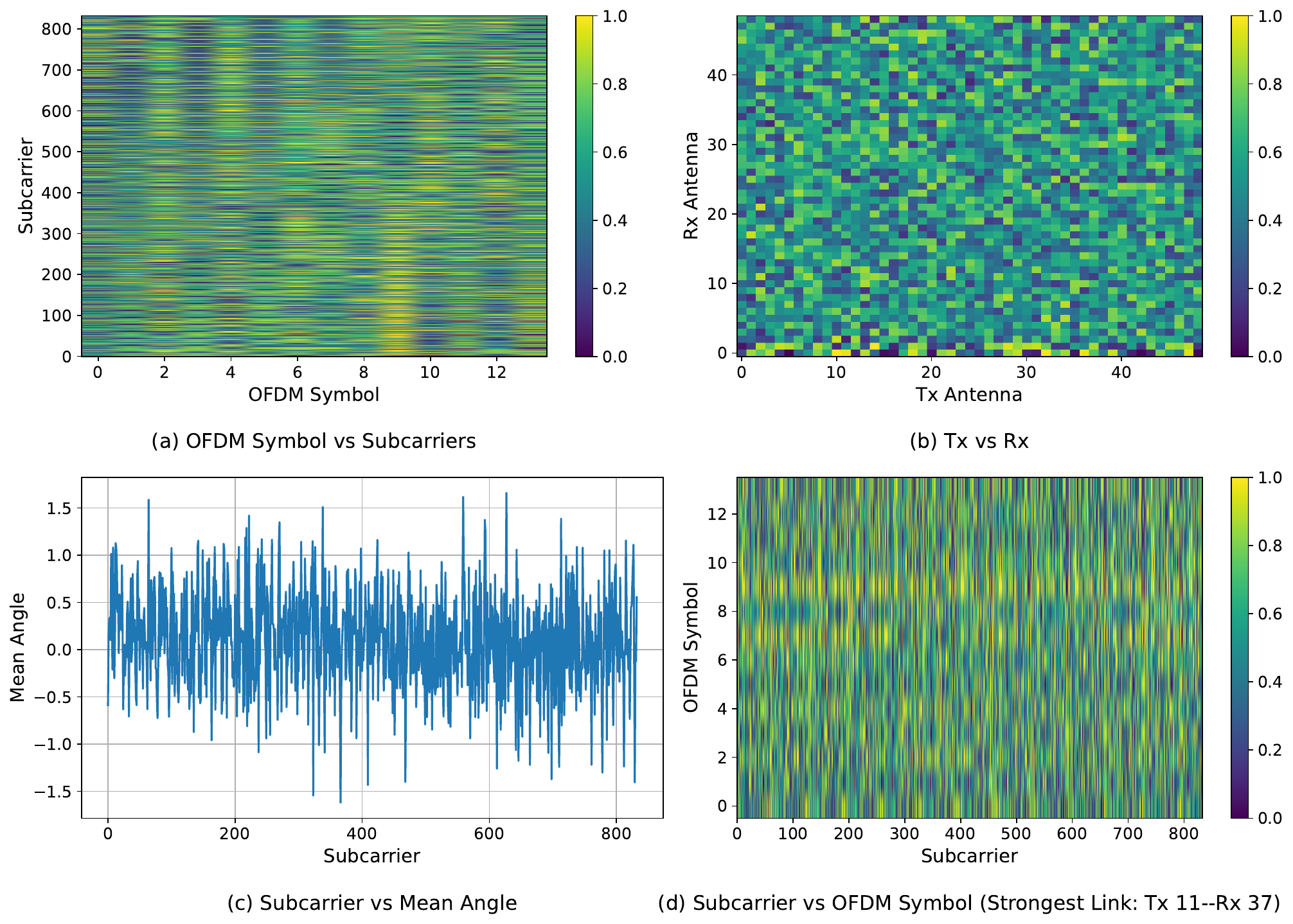}
    \caption{CSI phase visualization using the four-step method when an intruder is present in the danger zone. Current position of moving intruder is [-5.79, -10] m, and distance from intruder edge to train edge is 5.9 m. This complete visualization is plotted in a 2x2 grid, as described in the caption of Fig. \ref{fig:csi_mag}.}
    \label{fig:csi_phase}
\end{figure*}

The radio parameters were configured into two sets. Each set will be called a validation round, defined by specific parameters to test different operational scenarios. Validation round 1 is based on experimental trials with small-value radio parameters. In contrast, Validation round 2 is designed with a focus on practical deployment and theoretical accuracy. Both validation-round radio parameters are listed in Table \ref{table:radio_configuration}. The shapes of the CSI matrices obtained from validation round 1 are 100x100x30x1, and for validation round 2, they are 49x49x833x14. For simplicity in analysis and model training, we divided the dataset of both validation rounds into two classes. When the intruder is in the danger zone, we consider the CSI data class 1; when it is absent or in the safe zone, we consider it class 0. In validation round 1, a total of 11,295 CSI matrices are generated along with class, of which 6,186 belong to class 1, and the remaining 5,109 belong to class 0. Since validation round 2 parameters are practically appropriate, they were further analyzed using visual and statistical methods.
\begin{table}[!t]
\centering
\caption{Two sets of radio configurations are used for data generation, one for each validation round.}
\label{table:radio_configuration}
\begin{tabular}{lcc}
\toprule
\textbf{Parameter} & \textbf{Validation round 1} & \textbf{Validation round 2} \\
\midrule
Tx/Rx Array & 10x10 & 7x7\\
Antenna Spacing & $\lambda$ & $\lambda/2$\\
Frequency & 5 GHz & 5 GHz \\
Bandwidth & 120 KHz & 100 MHz  \\
Subcarrier Spacing & 4 KHz & 120 KHz  \\
Subcarriers & 30 & 833 \\
OFDM Symbols & 1 & 14  \\
\bottomrule
\end{tabular}
\end{table}
\subsubsection{Visualization of CSI matrix}
These CSI matrices are high-dimensional, contain complex numbers, and cannot be visualized as images or videos. There is no standard method of visualization, but we can do it in multiple steps. First, we only consider the magnitude of the CSI matrix and plot it in four steps.
\begin{itemize}
\item \textbf{Frequency-Time Profile}: This gives the overview of temporal views, and helps in motion detection of having shape 833x14. This profile can be evaluated by averaging along with the Tx and Rx dimensions.
\begin{figure}[!h] 
    \centering 
    \includegraphics[width=\linewidth, height=0.4\textheight, keepaspectratio]{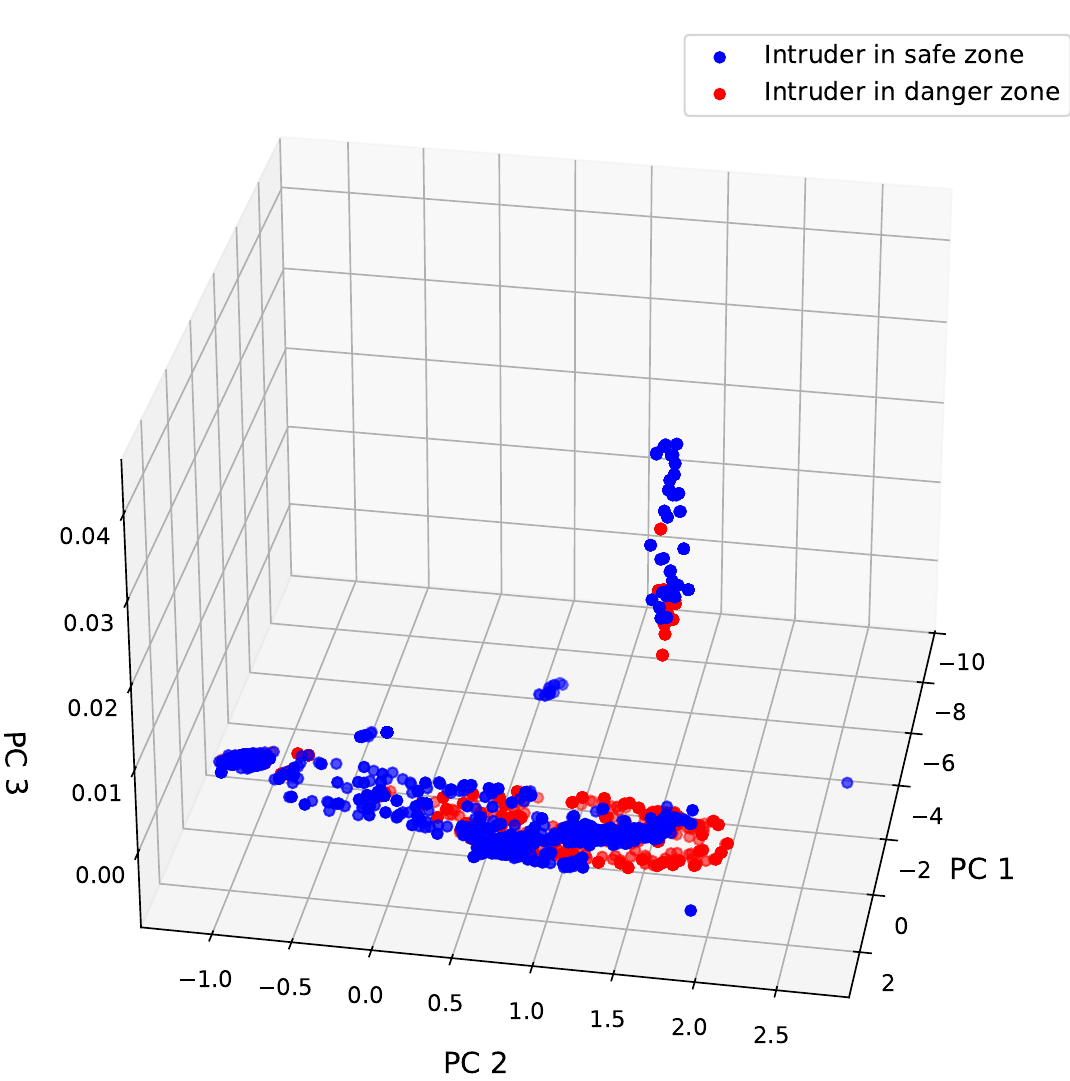} 
    \caption{Scattering plot of 1000 CSI data after application of PCA with 3 PCs. From each class, 500 CSI matrices are taken along with their class information. All the PCs are on axes of this 3D graph. Each tiny bubble represents a reduced-dimension CSI matrix, and red bubbles represent class 1 data, while blue bubbles represent class 0 data.} 
    \label{fig:pca_plot} 
\end{figure}
\item \textbf{Multiple-input and multiple-output(MIMO) Channel Structure}: This provides an antenna view and is represented as a single heatmap of shape 49x49, obtained by averaging along with OFDM, and subcarriers.
\item \textbf{Signal Evolution}: This gives a complete overview of the subcarriers' strength and helps in determining important frequencies.
\item \textbf{Strongest MIMO link}: It tells the most important path between the transmitter and receiver through the heatmap.
\end{itemize}
To depict the magnitude output by this method, we have used a CSI record when an intruder is present in the danger zone. The initial position of the intruder was [-8, -10] m; the current position of the intruder when CSI was recorded was [-5.79, -10] m; the distance from the train (edge-to-edge of objects) was 5.9 m; and the distance from the track was 5.79 m. From this distance value, we can see that the intruder is present in the danger zone, and the corresponding visual plot is given in Figure \ref{fig:csi_mag}. In subplot (a), we observe significant disturbances in the lower and higher subcarriers at some initial and end symbols, whereas the middle section shows almost flat magnitude. Although the plot for the safe zone, CSI,  is not shown here, it also shows similar behavior. In subplot (b), the MIMO channels exhibit a checkerboard-like structure due to the grid of antenna pairs. Each tile represents channel strength from Tx$(i)$ to Rx$(j)$. The subplot (c) also conveys the same message as subplot (a). Subplot (d) represent the strongest link between the transmitter array and the receiver array.
Now we will plot the CSI phase into the same four steps for the same CSI data. The output for the phase plot is given in Figure \ref{fig:csi_phase}. A significant amount of phase variation is observable, and therefore, taking the mean across symbols or frequency will yield noisy information. In subplots (a) and (d), we can see this behaviour. In subplot (b), the MIMO structure resembles a magnitude plot and appears coherent. The average magnitude of the subcarrier response values is quite good compared to the magnitude plot shown in subplot (c). Subplot (d) also indicates the same MIMO link as the strongest, as does the magnitude plot.

\subsubsection{Separability Analysis}
Since the dataset is divided into two classes, the separability of the CSI matrices across both classes provides insight into selecting the ML model. We have considered each CSI matrix and its corresponding class label as a pair of points in a different space. Principal Component Analysis (PCA) \cite{pca_paper} is a method for dimensionality reduction that can be used for class separation tests. PCA projects high-dimensional data onto orthogonal directions, called principal components (PC), that successively maximize the variance retained from the original data. We have taken 1000 data samples, 500 from each class, applied PCA with first three PCs, and plotted the results in Figure \ref{fig:pca_plot}.
From the scattered data points, we can observe that the data from both classes overlap so much, and visually, they are very less separable. This makes it clear that we need to pre-process the data in a good manner in order to extract suitable features from it, and a strong ML model to do classification and prediction.

\subsection{Data Pre-processing}
The input pipeline of the ML model uses multiple preprocessing methods to handle raw CSI files. We have pipelined the multiple preprocessing methods in a sequence. At each stage, a specific type of signal processing is applied to the CSI matrices. Below are the stages to preprocess the raw CSI data. 
\subsubsection{Remove Pilot and Null subcarriers}
Pilot and Null subcarriers do not carry any user data and are reserved for channel estimation and guard. Removing them can reduce the effective dimensionality of CSI matrices \cite{null, null2}. Removing Pilot subcarriers can slightly improve apparent spectral efficiency in post-processing, but prior knowledge of their indices is required.
\subsubsection{Signal to Noise Ratio (SNR)-weighted subcarrier selection}
In our CSI matrices, numerous subcarriers are used, but not all of them carry significant information. Because different subcarriers experience different fading, environmental effects, and noise, investigating their SNR can be very helpful in determining which subcarriers are important \cite{snr}. With the movement of the train and the intruder, there should be some change in the CSI values with frequencies. We have used a multi-criteria subcarrier selection approach that combines signal quality with discriminative power. This method computes a composite score for each subcarrier as $score[k] = SNR[k] * variance[k]$, using SNR and variance to identify strong and sensitive subcarriers. To reduce redundancy and enhance feature diversity, we incorporate de-correlation based on coherence bandwidth analysis. The coherence bandwidth, defined as $Bc = 0.5/\tau$, where $\tau$ is the Root Mean Square delay spreads, determines the subcarrier spacing threshold. Within each coherence band, we select the subcarrier with the highest SNR-weighted variance score, ensuring that selected features are both statistically independent and informative. The final subcarrier set is obtained by combining representatives from each de-correlated band and supplementing with additional high-scoring subcarriers from remaining bands until the target keep-ratio is achieved. This hybrid approach effectively balances signal strength, dynamic sensitivity, and feature independence to create robust feature representations for ML-based CSI classification and prediction.
\subsubsection{Static component removal}
Static component removal removes the constant part of the wireless channel that remains unchanged over time. It suppresses reflections from stationary objects like walls, ground, and infrastructure. This highlights only the dynamic variations caused by moving targets, such as intruders or vehicles. 
\begin{figure*}
    \centering
    \includegraphics[width=1\linewidth, height=1\textheight, keepaspectratio]{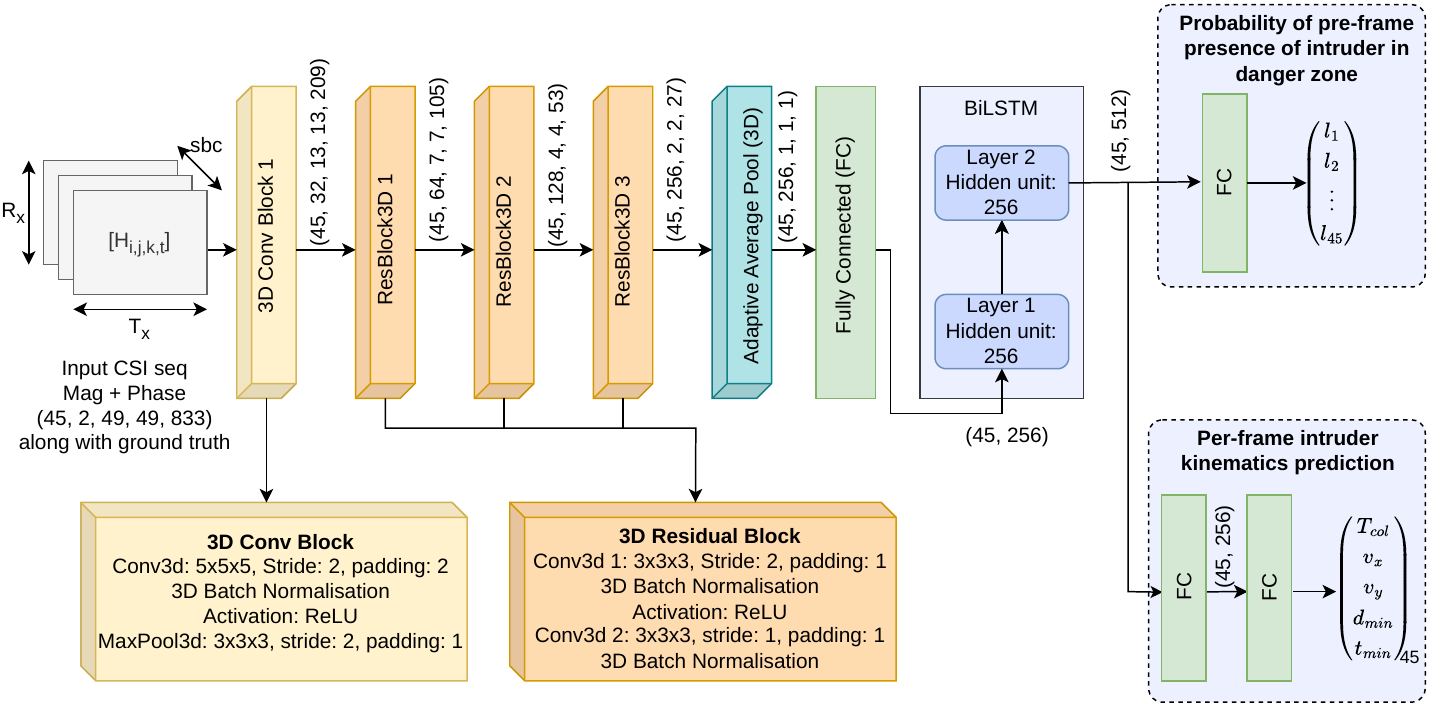}
    \caption{This is the proposed ML model architecture. It has two major sections: the first extracts spatial-temporal features, and the second is task-specific, performing classification and prediction of the intruder. It takes a sequence of CSI matrices as input (in the picture, we have used 45) and returns binary logits from the classification head and intruder kinematics from the regression head. The feature extractor contains a 3D Convolution block,  three residual 3D convolution blocks with skip connection, an adaptive average pooling layer, a fully connected layer, and a two-layer BiLSTM network. Details of the 3D convolutional blocks are shown in the boxes below. To generate the output classification and regression heads, use fully connected layers.}
    \label{fig:archi}
\end{figure*}
\subsubsection{Temporal denoising and average}
The Wiener filter is a linear filter that minimizes the mean-squared error (MSE) between the true channel and the filtered estimate, using the statistical properties of the signal and noise. The filter coefficients are calculated as $W = Rss / (Rss + Rnn)$, where $Rss$ is the signal autocorrelation and $Rnn$ is the noise autocorrelation. CSI temporal smoothing effectively suppresses random noise while preserving the non-stationary dynamics of wireless channels, including rapid fluctuations caused by intruder motion. The filter requires prior knowledge of signal and noise power spectral densities, which are estimated from CSI measurements. Additionally, we compute the mean across OFDM symbols to make the features more robust and less susceptible to noise.

After these four preprocessing pipelines, each complex-valued CSI matrix is reduced to a three-dimensional shape. Onward, it is separated into two real-valued components: Magnitude and Phase, and they are stacked along the zeroth axis to form a 4D tensor. Now these clean and processed CSIs are fed to the ML model.

\subsection{Intruder Detection and Kinematic Prediction ML Model}
We propose a two-stage deep learning-based architecture that combines spatial feature extraction from individual CSI frames with temporal feature extraction from CSI sequences. The model is specifically designed to predict intruder information based on a sequence of CSI matrices, where each CSI frame captures the spatial distribution of wireless channel properties. The architecture consists of two primary components: a \textit{FrameEncoder} that extracts 3D convolutional features from individual CSI matrices, and a \textit{SequenceEncoder} that aggregates frame-level features temporally using  BiLSTM layers to capture motion patterns and channel dynamics. The model architecture can be divided into two major sections as shown in Figure \ref{fig:archi}. The first section, which extracts features, consists of 3D Convolution and BiLSTM layers, and the second section predicts the outputs. The feature extractor has two very important Python classes, as mentioned below.

\textbf{Spatial Feature Extraction (FrameEncoder)}: The FrameEncoder is a 3D convolutional neural network that processes individual CSI frames represented as four-dimensional tensors of shape (C, H, W, D) where H and W are spatial dimensions, D is subcarrier depth, and C represents the two-channel input (magnitude and phase). The encoder performs progressive spatial down-sampling and depth incrementation through four convolutional blocks: (2→32 channels), (32→64 channels), (64→128 channels), and (128→256 channels). Except for the first convolution block, the remaining three blocks use two 3D convolution layers, each with a skip connection to ensure gradient flow. Following the convolutional layers, adaptive average pooling reduces the spatial dimensions to (256, 1, 1, 1), flattening the 256-dimensional feature maps. At last, a fully connected layer projects these features to 256 dimensions.
\begin{table*}
\centering
\caption{All the layers of the model and the corresponding output shapes for input shape (45, 2, 49, 49, 833)}
\label{tab:overall_architecture}
\resizebox{\textwidth}{!}{
\begin{tabular}{l|l|l|l|l|l|l}
\toprule
\textbf{Block} & \textbf{Layer} & \textbf{Input Shape} & \textbf{Output Shape} & \textbf{Kernel \& Stride} & \textbf{BN} & \textbf{Activation}\\
\midrule
\midrule
\multicolumn{7}{c}{\textbf{FrameEncoder}} \\
\midrule
\multirow{2}{*}{3D Conv Block 1} & 3D Convolution & $(45, 2, 49, 49, 833)$ & $(45, 32, 25, 25, 417)$ & $5\times5\times5$, stride=2 & $\checkmark$ & ReLU\\
 & 3D MaxPool & $(45, 32, 25, 25, 417)$ & $(45, 32, 13, 13, 209)$ & $3\times3\times3$, stride=2 & $-$ & $-$ \\
\midrule
\multirow{2}{*} {ResBlock3D 1} & 3D Convolution & $(45, 32, 13, 13, 209)$ & $(45, 64, 7, 7, 105)$ & $3\times3\times3$, stride=2 & $\checkmark$ & ReLU \\
& 3D Convolution & $(45, 32, 13, 13, 209)$ & $(45, 64, 7, 7, 105)$ & $3\times3\times3$, stride=1 & $\checkmark$ & $-$ \\
\midrule
\multirow{2}{*}{ResBlock3D 2} & 3D Convolution  & $(45, 64, 7, 7, 105)$ & $(45, 128, 4, 4, 53)$ & $3\times3\times3$, stride=2 & $\checkmark$ & ReLU\\
& 3D Convolution  & $(45, 64, 7, 7, 105)$ & $(45, 128, 4, 4, 53)$ & $3\times3\times3$, stride=1 & $\checkmark$ & $-$\\
\midrule
\multirow{2}{*}{ResBlock3D 3} & 3D Convolution  & $(45, 128, 4, 4, 53)$ & $(45, 256, 2, 2, 27)$ & $3\times3\times3$, stride=2 & $\checkmark$ & ReLU\\
& 3D Convolution  & $(45, 128, 4, 4, 53)$ & $(45, 256, 2, 2, 27)$ & $3\times3\times3$, stride=1 & $\checkmark$ & $-$\\
\midrule
AdaptiveAvgPool3D & Adaptive Pooling & $(45, 256, 2, 2, 27)$ & $(45, 256, 1, 1, 1)$ & $-$ & $-$ & $-$ \\
Linear & FC & $(45, 256)$ & $(45, 256)$ & $-$ & $-$ & $-$ \\
\midrule
\multicolumn{7}{c}{\textbf{SequenceEncoder}}\\
\midrule
\multirow{2}{*}{Temporal Block} & BiLSTM & $(45, 256)$ & $(45, 512)$ & $-$ & $-$ \\
 & BiLSTM & $(45, 512)$ & $(45, 512)$ & $-$ & $-$ \\
\midrule
\multicolumn{7}{c}{\textbf{Classification Head}} \\
\midrule
Linear & FC & $(45, 512)$ & $(45, 1)$ & $-$ & $-$ & Sigmoid \\
\midrule
\multicolumn{7}{c}{\textbf{Regression Head}} \\
\midrule
\multirow{2}{*}{Linear} & FC & $(45, 512)$ & $(45, 256)$ & $-$ & $-$ & ReLU\\
 & FC & $(45, 256)$ & $(45, 5)$ & $ $ & $-$ & $-$ \\
\bottomrule
\end{tabular}
}
\end{table*}

\textbf{Temporal Modelling (SequenceEncoder)}: The SequenceEncoder processes a sequence of $N$ CSI frames by first extracting per-frame features using the FrameEncoder, then modelling temporal dependencies through BiLSTM layers. The sequential feature has an input shape of (N, 256) and is passed to a two-layer BiLSTM with 256 hidden units. The BiLSTM captures temporal patterns and intruder motion signatures across the N-frame sequence.

This hybrid architecture leverages two complementary learning mechanisms: (1) FrameEncoder excels at learning local feature patterns from the 2D antenna grid and frequency-domain structure of CSI, and (2) BiLSTMs are utilized for capturing sequential dependencies and temporal information important for detecting motion signatures of an intruder. Through spatial and temporal learning, the model can learn interpretable representations: frame-level features capture snapshot channel properties (e.g., reflections, scattering patterns), while BiLSTM layers model how these properties vary as the target moves. Since all preprocessed CSI matrices are 3-dimensional but may differ in the number of subcarriers, we cannot use a fixed size. The input shape (45,2,49,49,833), the corresponding output dimensions of each layer, and the sequence of operations performed within the model's layers are presented in Table \ref{tab:overall_architecture}.

\subsection{Training of ML Model}
To promote stable convergence and prevent overfitting, the model is trained using a tuned set of hyper-parameters and hardware configurations. We optimized most of the model's hyper-parameters using Optuna \cite{optuna}, an automated hyperparameter optimization framework. The specific parameters, along with their respective optimized value and sampling methods, are detailed in Table \ref{tab:optuna}.

\begin{table}[!h]
\centering
\caption{Various hyper-parameters used in the model, their optimized values, and the optimization methods.}
\label{tab:optuna}
\begin{tabular}{lcc} 
\toprule
\textbf{Parameter} & \textbf{Optimum Value} & \textbf{Method}\\
\midrule
Learning Rate & $1.1 \times 10^{-5}$ & Tree Structured Parzen Estimation \\
Batch Size &$32$ & Bayesian Search  \\
Optimizer &$RMSProp$ & Bayesian search\\
Dropout &$0.1$ & Bayesian search\\
\bottomrule
\end{tabular}
\end{table}
To quantify the notion of a mistake, two loss functions are used to train the model. For the classification head, Binary Cross-Entropy (BCE) with Logits Loss is used; for the regression head, MSE Loss is used. The final loss is the accumulation of both losses, using the fact that backpropagation is performed. The model is trained for 100 epochs on a high-performance NVIDIA RTX A6000 GPU to maximize computational throughput. Although the training limit was set to 100, the model reached convergence at the 89th epoch.

\section{Experimental Validation}
In validation round 1, 11,295 CSI matrices and their corresponding ground-truth labels were generated. To ensure a rigorous validation process and prevent information leakage between the training and test sets, the dataset was partitioned simulation-wise using a hold-out strategy. All consecutive CSI matrices generated within a single simulation were kept together and assigned exclusively to either the training or test set. Thus, CSI matrices originating from the same simulation trajectory were never distributed across both sets. The order of CSI matrices within each simulation was preserved to maintain their temporal dependencies. Following this strategy, 2,259 CSI matrices were assigned to the test set, while the remaining 9,036 CSI matrices were used for model training. Furthermore, the architecture's temporal sensitivity was evaluated by training the model with varying CSI sequence lengths, thereby assessing its ability to capture dynamic environmental changes over time. To investigate the robustness of the model, we will examine classification and regression results separately, and finally, in Inference, together.
\subsection{Intruder Presence Detection} 
The detection efficacy of the model is quantified using standard classification metrics, Accuracy, Precision, Recall, and the F1-score. The performance metrics for the validation rounds on test data with varying sequence length are presented in Tables \ref{tab:metrics_results1} and \ref{tab:performance_results_2}, respectively.
\begin{table}[!t]
    \centering
    \caption{Intruder presence classification performance of the model with different CSI sequence lengths on validation round 1 test data.}
    \label{tab:metrics_results1}
    \begin{tabular}{ccccc}
        \toprule
        \textbf{Sequence length} & \textbf{Accuracy} & \textbf{Precision} & \textbf{Recall} & \textbf{F1 Score} \\
        \midrule
        1   & 0.9000 & 0.8801 & 0.8700 & 0.8740 \\
        10  & 0.9737 & 0.9710 & 0.9563 & 0.9636 \\
        45  & 0.9990 & 0.9992 & 0.9982 & 0.9987 \\
        120 & 0.9999 & 0.9999 & 0.9997 & 0.9998 \\
        \bottomrule
    \end{tabular}
\end{table}
Both table results indicate a positive correlation between input sequence length and overall model performance. Performance increases with the length of the input sequence. This improvement indicates the model's capability to extract deeper temporal relationships from the data as more historical CSI matrices are provided. Since the simulations are carried out in a sequence of steps, the steps are related to one another. Increasing the sequence length allows the architecture to achieve precise, robust decision-making and be less prone to false alarms than "greedy" instantaneous prediction. While a sequence length of 1 results in high variance and decision errors, longer sequences enable the model to analyze the intruder's trajectory over time precisely. However, a trade-off exists in sequence length and operational latency; for instance, a sequence length of 120 introduces significant processing delays that could result in late intruder detection. Sequence lengths of 10 and 45 emerge as the practical configurations, having an optimal balance between predictive accuracy and the low-latency response times required for railway safety applications.

The Validation round 2 is an important phase of the model evaluation, as the radio parameters were set to align with real-world standards. For this configuration, a dataset of size 11,400 samples was generated through multiple simulations. The dataset was partitioned simulation-wise using an 80:20 train-test split, resulting in 9,120 CSI matrices for training and 2,280 for testing. As in Validation Round 1, all CSI matrices belonging to a given simulation were kept within the same partition, while their temporal order was preserved. By applying realistic configurations such as proper subcarrier spacing and large bandwidth, this configuration provides a high-fidelity assessment of the system's sensing capabilities. The predictive performance of the model under these conditions is summarized in Table \ref{tab:performance_results_2}, which details the accuracy and reliability. From these two results, we can observe that performance across sequence length is predictable regardless of radio configuration. The model is learning the classification task without overfitting on large shape CSI data. It also shows persistent behaviors across both validation rounds, which indicates the model's robustness.
\begin{table}[!t]
    \centering
    \caption{Intruder presence classification performance for different CSI sequence lengths on the Validation Round 2 test dataset.}
    \label{tab:performance_results_2}
    \begin{tabular}{ccccc}
        \toprule
        \textbf{Sequence length} & \textbf{Accuracy} & \textbf{Precision} & \textbf{Recall} & \textbf{F1 Score} \\
        \midrule
        1   & 0.9271 & 0.9310 & 0.9152 & 0.9231 \\
        10  & 0.9584 & 0.9200 & 0.9534 & 0.9364 \\
        45  & 0.9654 & 0.9732 & 0.9381 & 0.9553 \\
        120 & 0.9957 & 0.9955 & 0.9932 & 0.9944 \\
        \bottomrule
    \end{tabular}
\end{table}
\subsection{Intruder Kinematic Prediction}
The kinematic prediction performance is evaluated using Mean Absolute Error (MAE) and Mean Squared Error (MSE). The regression head predicts the relative position and velocity of the intruder with respect to the train, together with collision-related parameters such as time-to-collision and minimum separation distance. Since relative position is a fundamental quantity for assessing the intruder–train trajectory, we primarily report its prediction error in Table \ref{tab:mae}. On the validation round 1 test dataset, an MAE of 0.4200 indicates that the model is making mistakes of only 0.4200 meters in relative position prediction. This shows the model's strong ability to predict relative positions.
\begin{table}[!h]
\centering
\caption{A summary of errors made in the prediction of relative position on the training and test data for validation round 1.}
\label{tab:mae}
\begin{tabular}{lcc}
\toprule
 & \textbf{MAE} & \textbf{MSE} \\
\midrule
Train & 0.2837 & 0.4737 \\
Test  & 0.4200 & 0.7042 \\
\bottomrule
\end{tabular}
\end{table}
\begin{figure}[!h]
    \centering
    \includegraphics[width=1\linewidth]{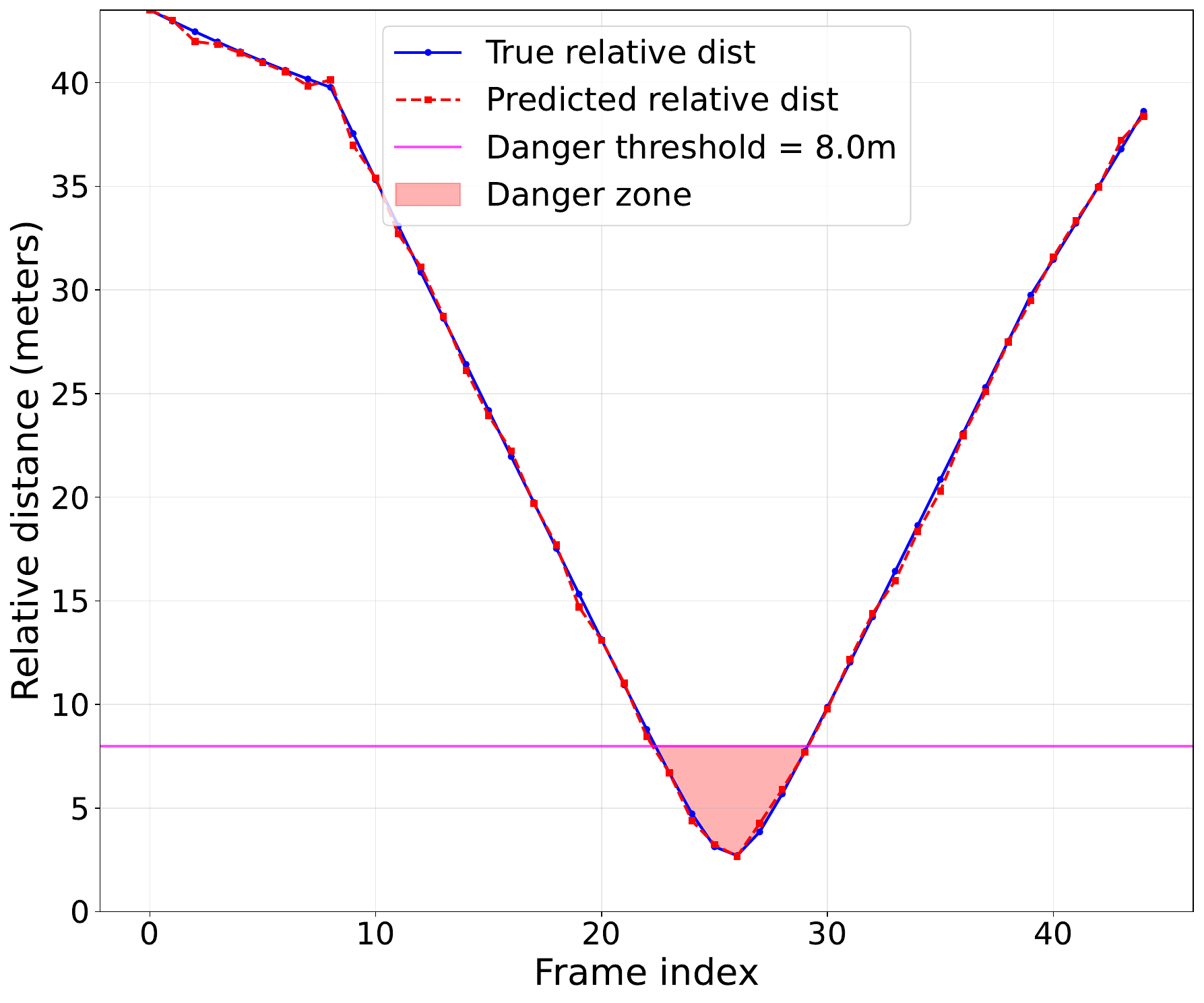}
    \caption{Predicted trajectory of the relative position of the intruder with respect to the train, given with frame index or time on 45 CSI matrices obtained from one simulation. The blue color line shows ground truth, the red color dashed line shows the predicted value, and the magenta line is the threshold line. In this case, the threshold value is obtained by adding 3 meters to the half dimension of the train and the intruder (Since this measurement is taken from the center of each object).}
    \label{fig:trajectory}
\end{figure}
\subsection{Model Inference}
To provide a qualitative assessment of model inference, we examine a representative sequence of 45 CSI matrices along with ground truth from the test data, to verify the model's predictions. We have taken a simulation's data in which the intruder started from the initial position [-20, 7] m with a velocity of 1 m/sec towards the positive x-axis. The ground truth and the model's prediction are given in Table \ref{tab:inference}, and the relative position of the intruder with respect to the train is shown in Figure \ref{fig:trajectory}. This table shows that the model is predicting Velocity, time of collision, and minimum distance of approach precisely with very little error.
\begin{table}[!h]
\centering
\caption{Intruder kinematic predictions over a sequence of 45 CSI matrices, with ground truth and corresponding errors.}
\label{tab:inference}
\begin{tabular}{lllc}
\toprule
\textbf{Parameter} & \textbf{Prediction} & \textbf{Ground Truth} & \textbf{Error} \\
\midrule
Chance of Collision & Yes & Yes & -- \\
Velocity ($V_x$) & 0.999 m/s & 1 m/s & 0.001 \\
Velocity ($V_y$) & -0.012 m/s & 0 m/s & 0.012 \\
Time to Collision & 4.618 s & 4.800 s & 0.182 \\
Minimum Distance & 2.603 m & 2.707 m & 0.104 \\
\bottomrule
\end{tabular}
\end{table}
In the predicted trajectory, the blue line represents the true distance between the intruder and the train, and the red dashed line is the predicted distance. Between frame indices 5 and 10, the curve starts moving rapidly downward because the train was briefly stopped and then resumed moving. Here, the threshold is 8 meters instead of 3 because the half dimension of the train and the intruder are both considered. It is evident that between frame indices 20 and 30, the relative distance falls below the threshold, indicating that an accident will occur. 

This matches the ground truth, and the region shaded in red shows CSIs that are under the danger zone criterion. Overall, the inference results demonstrate that the proposed model effectively captures the temporal dynamics of the intruder–train interaction from CSI measurements. 

\section{Conclusion and limitations}
This work demonstrates the potential of CSI-based 6G sensing for intelligent railway intrusion detection and collision-risk prediction. The presented results show that, with appropriate radio configurations, CSI preprocessing, and machine learning techniques, variations in the wireless channel can be effectively exploited to detect intruders and estimate their motion-related parameters, including position and velocity. The proposed 3D CNN--BiLSTM model further demonstrates that the spatial and temporal characteristics of CSI can be jointly learned to identify intrusion events and predict potential collision scenarios. These findings highlight the capability of CSI to serve as a valuable sensing information source for future ISAC-enabled railway safety applications. With suitable radio resources, computational infrastructure, and real-world validation, the proposed approach has the potential to support real-time railway monitoring and proactive collision-risk assessment.

Several technical factors have been identified that influence the performance of CSI-based railway sensing. First, intrusion detection and localization performance are strongly dependent on the MIMO configuration and transmit power. As the antenna grid size decreases, performance decreases rapidly. Second, the physical dimensions and material properties of the intruder affect its behavior with the wireless signal; while larger objects generally produce more distinguishable CSI variations, detecting smaller or highly absorptive objects remains challenging. Third, effective motion characterization relies on sufficient temporal variation in CSI. Slowly moving or stationary intruders may produce weaker temporal signatures, making velocity and trajectory estimation more difficult. Furthermore, scenarios involving variable intruder velocities and complex trajectory properties require more comprehensive study. Finally, the current study is based on synthetically generated CSI under simulated propagation conditions; therefore, validation using real-world railway measurements is necessary to assess the robustness and generalizability of the proposed framework before practical deployment.


\begin{thebibliography}{00}
\bibitem{sionna} Hoydis, Jakob, et al. "Sionna: An open-source library for next-generation physical layer research." arXiv preprint arXiv:2203.11854 (2022).
\bibitem{blender} https://www.blender.org/
\bibitem{csi} Foschini, G. J. (1996). Layered space-time architecture for wireless communication in a fading environment when using multi-element antennas. Bell labs technical journal, 1(2), 41-59.
\bibitem{rail_safe} https://railsafe.org.au/\_media/documents/resources/worksite-protection/Rail-Industry-Safety-Induction-RISI-Handbook.pdf
\bibitem{wifi_vision} He, Ying, et al. "WiFi vision: Sensing, recognition, and detection with commodity MIMO-OFDM WiFi." IEEE Internet of Things Journal 7.9 (2020): 8296-8317.
\bibitem{elephant} Suggested Measures to Mitigate Elephant \& Other Wildlife Train Collisions on Vulnerable Railway Stretches in India
\bibitem{isac_uses} Elnashar, Ayman, Marwan Bin Shakar, and Sami Muhaidat. "Integrated Sensing and Communication in 6G: A Comprehensive Survey of Use Cases, Enabling Technologies, Standardization Roadmap, and Monetization Frameworks." IEEE Open Journal of the Communications Society (2026).
\bibitem{ofdm} Chang, R. W. (1966). Synthesis of band‐limited orthogonal signals for multichannel data transmission. Bell system technical journal, 45(10), 1775-1796.
\bibitem{wireless_sensing} Ge, Yao, et al. "Contactless WiFi sensing and monitoring for future healthcare-emerging trends, challenges, and opportunities." IEEE Reviews in Biomedical Engineering 16 (2022): 171-191.
\bibitem{pca_paper} Pearson, Karl. "LIII. On lines and planes of closest fit to systems of points in space." The London, Edinburgh, and Dublin philosophical magazine and journal of science 2.11 (1901): 559-572.
\bibitem{lstm} Hochreiter, Sepp, and Jürgen Schmidhuber. "Long short-term memory." Neural computation 9.8 (1997): 1735-1780.
\bibitem{bayesteh} Bayesteh, Alireza, et al. "Integrated sensing and communication (ISAC)—From concept to practice." Communications of Huawei Research (2022): 4-25.
\bibitem{ghosh} Ghosh, Amitava, et al. "A unified future: Integrated sensing and communication (ISAC) in 6G." IEEE Journal of Selected Topics in Electromagnetics, Antennas and Propagation (2025).

\bibitem{isac_csi} D. Zhang, D. Wu, K. Niu, X. Wang, F. Zhang, J. Yao, D. Jiang, and F. Qin, “Practical Issues and Challenges in CSI-based Integrated Sensing and Communication,” 2022.

\bibitem{sebastian} Robitzsch, Sebastian, et al. "Architecture Considerations for ISAC in 6G." 2025 IEEE Conference on Standards for Communications and Networking (CSCN). IEEE, 2025.
\bibitem{LiDAR} Nan, Zongliang, et al. "A novel high-precision railway obstacle detection algorithm based on 3D LiDAR." Sensors 24.10 (2024): 3148.
\bibitem{Shanping} Ning, Shanping, et al. "Railway Intrusion Risk Quantification with Track Semantic Segmentation and Spatiotemporal Features." Sensors 25.17 (2025): 5266.

\bibitem{Hu} Hu, Tingyi, Feng Gao, and Fuqiang Zhou. "Railway obstacle intrusion detection and risk assessment based on MSIA-YOLOv8 and DALNet." Expert Systems with Applications (2025): 130132.
\bibitem{3gpp} 3rd Generation Partnership Project; Technical Specification Group TSG SA; Feasibility Study on Integrated Sensing and Communication (Release 19).
\bibitem{challenges} Sanz Bobi, Juan de Dios, et al. "Prediction of degraded infrastructure conditions for railway operation." Sensors 24.8 (2024): 2456.
\bibitem{optuna} Akiba, Takuya, et al. "Optuna: A next-generation hyperparameter optimization framework." Proceedings of the 25th ACM SIGKDD international conference on knowledge discovery \& data mining. 2019.
\bibitem{3D_LiDAR} Zhao, Anxin, Yekai Zhao, and Qiuhong Zheng. "Robust Pedestrian Detection and Intrusion Judgment in Coal Yard Hazard Areas via 3D LiDAR-Based Deep Learning." Sensors 25.18 (2025): 5908.
\bibitem{MACENet} Chen, Xichun, et al. "Automatic detection of foreign object intrusion along railway tracks based on MACENet." Plos one 20.8 (2025): e0329303.
\bibitem{Animal_detection} Dorji, Tsheten, et al. "AI-based Animal Intrusion Detection System for Human-Wildlife Conflicts in Bhutan." Zorig Melong| A Technical Journal of Science, Engineering and Technology 8.1 (2025): 46-52.
\bibitem{BCD-YOLO} Wang, Xiaopeng, Ce Han, and Weidong Jin. "BCD-YOLO: a railway perimeter foreign body intrusion detection method based on Yolov8." Measurement Science and Technology 36.5 (2025): 056007.
\bibitem{BriGuard} Liu, L., Zhang, W., Deng, C., Yin, S. and Wei, S. (2015), BriGuard: a lightweight indoor intrusion detection system based on infrared light spot displacement. IET Sci. Meas. Technol., 9: 306-314. https://doi.org/10.1049/iet-smt.2013.0171

\bibitem{positioning} Wymeersch, Henk, et al. "6G positioning and sensing through the lens of sustainability, inclusiveness, and trustworthiness." IEEE Wireless Communications 32.1 (2025): 68-75.
\bibitem{ray_trace} Hoydis, Jakob, et al. "Sionna RT: Differentiable ray tracing for radio propagation modeling." 2023 IEEE Globecom Workshops (GC Wkshps). IEEE, 2023.
\bibitem{null} Moshiri, Parisa Fard, et al. "A CSI-based human activity recognition using deep learning." Sensors 21.21 (2021): 7225.

\bibitem{null2} Koo, Minseok, and Jaesung Park. "KAN-Sense: Keypad input recognition via CSI feature clustering and KAN-based classifier." Electronics 14.15 (2025): 2965.
\bibitem{snr} Wang, Ruilin, et al. "A Subcarrier Selection Method for Wi-Fi-based Respiration Monitoring using IEEE 802.11 ac/ax Protocols." 2022 IEEE MTT-S International Microwave Biomedical Conference (IMBioC). IEEE, 2022.

\bibitem{isac} Liu, Fan, et al. "Integrated sensing and communications: Toward dual-functional wireless networks for 6G and beyond." IEEE journal on selected areas in communications 40.6 (2022): 1728-1767.
\bibitem{3d_cnn} Ji, Shuiwang, et al. "3D convolutional neural networks for human action recognition." IEEE transactions on pattern analysis and machine intelligence 35.1 (2012): 221-231.
\bibitem{csi_ml} Ohtsuki, Tomoaki. "Machine learning in 6G wireless communications." IEICE Transactions on Communications 106.2 (2023): 75-83.

\end{thebibliography}
\end{document}